\documentclass[10pt,twocolumn,letterpaper]{article}

\usepackage{wacv}              

\definecolor{wacvblue}{rgb}{0.21,0.49,0.74}
\usepackage[pagebackref,breaklinks,colorlinks,allcolors=wacvblue]{hyperref}
\usepackage{color,soul}
\usepackage[table]{xcolor}
\usepackage{xcolor} 
\usepackage[normalem]{ulem}

\def\wacvPaperID{} 
\def\confName{WACV}
\def\confYear{2026}

\title{RangeSAM: On the Potential of Visual Foundation Models for Range-View represented LiDAR segmentation}

\author{Paul Julius Kühn 
\and
Duc Anh Nguyen
\and
Arjan Kuijper
\and 
Saptarshi Neil Sinha \\
\and 
{\tt\small first.second.lastname@igd.fraunhofer.de}
}

\begin{document}
\def\ie{i.e.\ }
\def\eg{e.g.\ }
\def\etal{et~al.\ }
\def\wrt{w.r.t.\ }


\newcommand{\todoo}[1]{\textcolor{red}{#1}}
\newcommand{\nrevision}[2]{{\color{red}\sout{#1}}{\color{blue}\uwave{#2}}}
\newcommand{\jkrevision}[2]{{\color{red}\sout{#1}}{\color{blue}\uwave{#2}}}
\newcommand{\finalrevision}[2]{{}{\color{black} #2}}
\newcommand{\mwrevision}[2]{{}{\color{black} #2}}
\maketitle
\begin{abstract}LiDAR point cloud segmentation is central to autonomous driving and 3D scene understanding. While voxel- and point-based methods dominate recent research due to their compatibility with deep architectures and ability to capture fine-n  geometry, they often incur high computational cost, irregular memory access, and limited runtime efficiency due to scaling issues. In contrast, range-view methods, though relatively underexplored - can leverage mature 2D semantic segmentation techniques for fast and accurate predictions. Motivated by the rapid progress in Visual Foundation Models (VFMs) for captioning, zero-shot recognition, and multimodal tasks, we investigate whether SAM2, the current state-of-the-art VFM for segmentation tasks, can serve as a strong backbone for LiDAR point cloud segmentation in the range view representations. We present \textbf{RangeSAM}, to our knowledge, the first range-view framework that adapts SAM2 to 3D segmentation, coupling efficient 2D feature extraction with projection/back-projection to operate on point clouds. To optimize SAM2 for range-view representations, we implement several architectural modifications to the encoder: (1) a novel \textbf{Stem} module that emphasizes horizontal spatial dependencies inherent in LiDAR range images, (2) a customized configuration of \textbf{Hiera Blocks} tailored to the geometric properties of spherical projections, and (3) an adapted \textbf{Window Attention} mechanism in the encoder backbone specifically designed to capture the unique spatial patterns and discontinuities present in range-view pseudo-images.
Our approach achieves competitive performance on SemanticKITTI while benefiting from the speed, scalability, and deployment simplicity of 2D-centric pipelines.
This work highlights the viability of VFMs as general-purpose backbones for point cloud segmentation and opens a path toward unified, foundation-model-driven LiDAR segmentation. Results let us conclude that range-view segmentation methods using VFMs lead to promising results. 
\end{abstract}
    
\section{Introduction}
\label{sec:intro}

\begin{figure*}[htb!]
\centering  \includegraphics[width=0.85\linewidth]{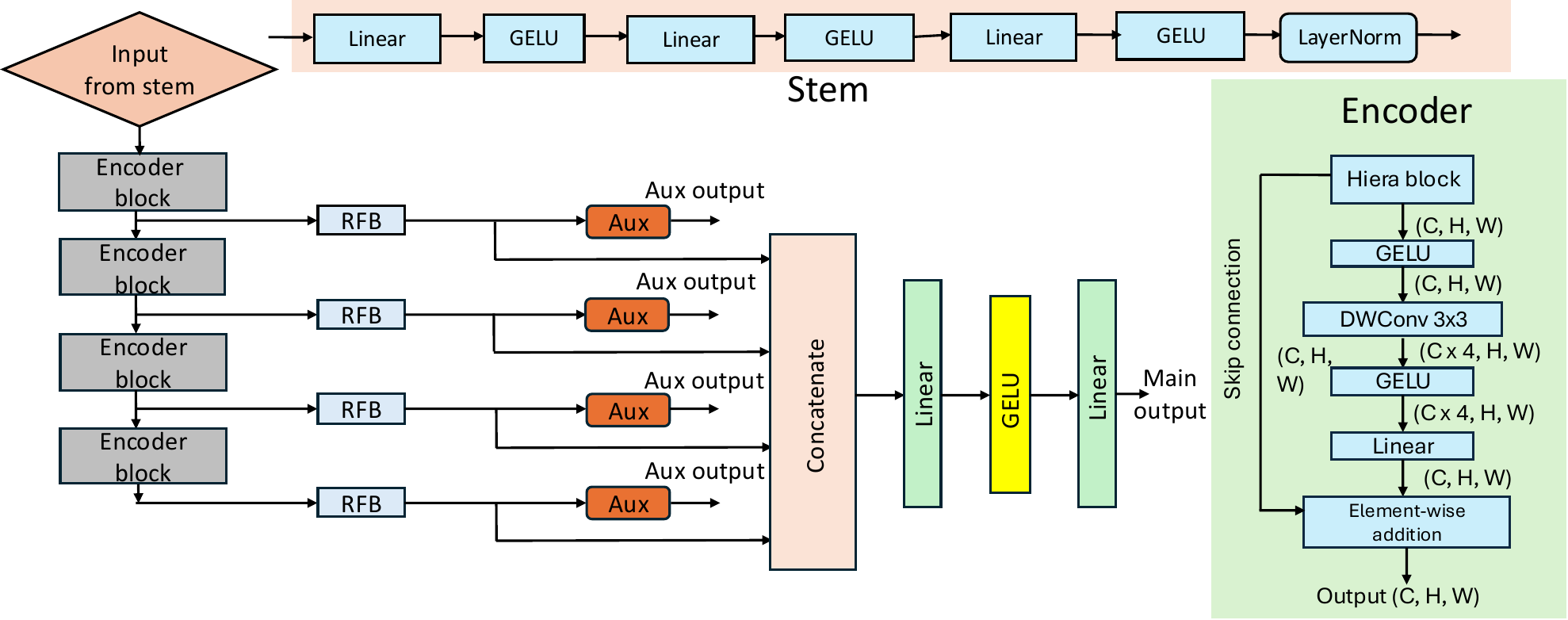}
  \caption{Overview of our SAM2-based point-cloud segmentation model. The stem reshapes range-view point clouds into a tensor suitable for the encoder. The encoder is built from stacked Hiera blocks—each containing a multi-head self-attention module and a feed-forward network~\cite{SAM2}. The decoder comprises of Receptive Field Blocks~\cite{ReceptiveFieldBlock} (RFB) with LayerNorm~\cite{layernormalization} and GELU~\cite{GELU}, concatenates multi-scale features, and projects them to $N_{\text{classes}}$ while also adding auxiliary head (Aux) on corresponding output.}
  \label{fig:pipeline_detail}
\end{figure*}

Reconstructing urban scenes with 2D and 3D semantics is crucial for autonomous systems (vehicles, drones, robots), requiring real-time viewpoint synthesis, semantic segmentation, depth generation, and dynamic object tracking. Among these, LiDAR point cloud semantic segmentation is fundamental for differentiating vehicles, pedestrians, road signs, and infrastructure.
Recent research focuses on voxel- and point-based methods \cite{ptv1, ptv2, ptv3, Cylinder3D, PVKD, SPVNAS}, which achieve strong performance but impose substantial computational and memory costs on large-scale outdoor data \cite{PointNetPlusPlus,PointConv, PointASNL, DynGraphCNN, graphattentionnetworks, JSIS3D, 3D-MPA, KPConvFA, PointNet} and struggle with irregular, unordered point clouds.
In contrast, range-view segmentation \cite{RangeFormer, RangeSeg, RangeViT, RangeNet++, FrustumPointNets} projects 3D point clouds into dense 2D representations, enabling reuse of mature 2D models with reduced memory and faster inference. Though previously overlooked due to limitations in handling occlusions and resolution loss, recent advances in attention mechanisms \cite{attention}, multi-scale fusion \cite{zhao2025multiscalefusionobjectrepresentation}, and context-aware architectures \cite{ContextAwareML} warrant reassessing the range-view paradigm.
We propose \emph{RangeSAM}, which adapts the Segment Anything Model 2 (SAM2) \cite{SAM2} for range-view LiDAR segmentation (Figure \ref{fig:pipeline_detail}). Our approach leverages SAM2's robust \emph{zero-shot capabilities} for 2D understanding and extends them to 3D via range-view representation. The methodology includes: (1) range projection preprocessing to transform unordered LiDAR scans, (2) a multi-component architecture incorporating Receptive Field Blocks~\cite{ReceptiveFieldBlock}, (3) postprocessing with \emph{k}-NN label propagation, and (4) a composite loss function addressing class imbalance and boundary accuracy. Our main contributions are:
\begin{itemize} \item We introduce RangeSAM, the first framework adapting SAM2 for LiDAR point cloud segmentation via range-view representations. \item We have designed a multi-component encoder architecture with a pretrained Hiera backbone, custom Stem module, novel embedding matrix, and Hiera blocks utilizing localized and global Multi-Head Attention (Figure~\ref{fig:pipeline_detail}). \item We demonstrate competitive performance on SemanticKITTI~\cite{semKitti1}, validating our approach's viability, and conduct ablation studies on training strategies and data augmentation. \end{itemize}

\section{Related Work}
\label{sec:rel_work}
\textbf{Visual Foundation Models}\\
Recent large-scale vision models have revolutionized image tasks \cite{florence1, florence2advancingunifiedrepresentation, groundingDino, OnTheOppsOfVFMs}. The Segment Anything Model (SAM) series achieved promptable segmentation with zero-shot transferability \cite{SAM1,samNetPlusPlus,GroundedSAM, semanticSAM}, while SAM2 extends to videos and complex scenarios \cite{SAM2,SAM2Adapter,sam2unet, huo2025sambaunetsynergizingsam2mamba,liu2025auralsam2enablingsam2hear, yuan2025sa2vamarryingsam2llava}, and has been adapted to 3D \cite{sam4d,sam3dsegment3dscenes, sam3dsegmentMedical}. Other VFMs include self-supervised transformers (DINO series \cite{DINOv1,DINOv2,DINOv3}), multimodal models \cite{OpenScene, seem}, and the EVA series \cite{EVA01,EVA-CLIP,EVA-CLIP-18B, EVA02}, offering strong generalization and label-efficient learning. For instance, DINOv2 features improve 3D semantic segmentation \cite{dinoroomleveraging2d, sam4d}, while CLIP's image-text embeddings enable zero-shot 3D understanding \cite{clip2scene, OpenScene,CLIPFO3D,ovoopenvocabularyoccupancy}.
\subsection{Methods on Projection-Based representations}
\textbf{Convolutional Neural Networks (CNNs):} Projection-based methods transform point clouds into structured 2D representations (range-, bird's-eye-, or perspective-view) to leverage mature 2D CNN architectures \cite{xu2020squeezesegv3, PerceptionAware, PolarNet, semKitti2, semKitti1, sieseffusionnetspatialintercorrelationenhancement, ScanBasedSeg, KPConvFA, LiteHDNet, RangeNet++, SuperpointGraphs, Kochanov2020KPRNetIP, RandLA, cortinhal2020salsanext}. The SqueezeSeg series \cite{wu2017squeezeseg, wu2018squeezesegv2, xu2020squeezesegv3, wang2019latte} pioneered real-time segmentation with lightweight CNNs and CRF post-processing \cite{CRF}. Methods like DarkNetSeg \cite{semKitti1}, SalsaNext \cite{cortinhal2020salsanext}, KPRNet \cite{Kochanov2020KPRNetIP}, RangeNet++ \cite{RangeNet++}, Lite-HDSeg \cite{LiteHDNet}, and SqueezeSegV3 \cite{xu2020squeezesegv3} employ a two-stage approach: range projection followed by U-Net-style architectures \cite{UNet}. RangeSeg works \cite{RangeSeg, MetaRangeSeg} incorporated multi-scale context modeling for spherical projections.\\
\textbf{Vision Transformers (ViTs):}
ViT architectures have driven significant advances: RangeViT and FRNet \cite{RangeViT, xu2025frnet} introduced hybrid CNN-Transformer backbones combining convolutional inductive biases with global attention, benefiting from 2D transfer learning \cite{Cityscapes}, while RangeFormer \cite{RangeFormer} established a fully transformer-based architecture achieving state-of-the-art performance on outdoor benchmarks \cite{semKitti1, semKitti2, nuScenes, waymo, semanticPOSS}.
\section{Methodology}
\label{sec:methodology}
We introduce \textbf{RangeSAM}, a projection-based LiDAR segmentation model leveraging VFMs. We detail the range projection preprocessing, architectural modifications, and postprocessing procedures. Figure~\ref{fig:pipeline_detail} illustrates the pipeline, including the stem module, SAM2-based encoder, and decoder components.
\subsection{Range Projection}
Each LiDAR scan is initially represented as an unordered point set $\mathcal{P} = {(x, y, z, f)}$, where $(x, y, z)$ denotes the Cartesian coordinates and $f$ represents auxiliary sensor measurements such as intensity or remission values. Subsequently, the point cloud undergoes transformation to a range view representation via projection onto the sensors spherical coordinate system \cite{RangeNet++}:
\begin{equation}
    \theta=\arctan \left(\frac{y}{x}\right), \phi=\arcsin \left(\frac{z}{r}\right),r=\sqrt{x^2+y^2+z^2}
\end{equation}
Specifically, the 3D points are discretized through rasterization into a 2D cylindrical projection with dimensions $H \times W$:
\begin{equation}
    \binom{u}{v}=\binom{\frac{1}{2}\left[1-\arctan (y, x) \pi^{-1}\right] w}{\left[1-\left(\arcsin \left(z r^{-1}\right)+\mathfrak{f}_{\mathrm{up}}\right) \mathfrak{f}^{-1}\right] h},
\end{equation}
where $(u, v)$ represent the projected coordinates of point $p_n$ in the range image coordinate system, and $h$ and $w$ denote the image height and width parameters, respectively. $r=\sqrt{x^2+y^2+z^2}$
denotes the range between a point and its sensor, the sensors vertical field of view $\mathfrak{f}$ comprises the sum of ($\mathfrak{f}_{\mathrm{up}}$ $+$ $\mathfrak{f}_{\mathrm{down}}$) viewing angles. For our method, we configured the imaging system to operate at the commonly used spatial resolution of $64 \times 2048$ pixels \cite{RangeViT, RangeSeg, RangeFormer}. Multiple points projecting to the same pixel were resolved by retaining the minimum-range feature. Unprojected pixels were zero-filled.

\subsection{Model Architecture}
SAM2 is a state-of-the-art VFM trained on the SA-V dataset \cite{SAM2}. SAM2-UNet \cite{sam2unet} achieves superior performance across multiple segmentation tasks while maintaining simplicity. RangeSAM adopts a similar paradigm (Figure~\ref{fig:pipeline_detail}), incorporating Receptive Field Blocks (RFB) \cite{ReceptiveFieldBlock} for enhanced feature decoding.
\textbf{Stem:} Transforms input tensors from (B,6,H,W) to (B,96,H,W) via linear transformation, layer normalization \cite{layernormalization}, and GELU activation \cite{GELU} (Figure~\ref{fig:pipeline_detail} top), then features were partitioned into overlapping 7×7 patches with unit stride. We replace SAM2's positional embedding with a novel (4,128) embedding matrix to enhance horizontal spatial sensitivity.
\textbf{Encoder:} Uses the pretrained Hiera \cite{Hiera, SAM2} backbone with a multi-component architecture (Figure~\ref{fig:pipeline_detail} right) comprising:
\textbf{Hiera block:} Each stage of the backbone consists of a customized number of Hiera blocks, and each block includes two modules: \textit{1) Multi-Head Attention module}, in early stages, attention is restricted within a window, while later stages may also use global attention. The operation can be written as
\begin{equation}
X_{out} = X + \text{DropPath}(\text{MHA}(\text{LayerNorm}(X)))
\end{equation}
\begin{equation}
\text{MHA}(Q, K, V) = \text{Concat}(\text{head}_1, \ldots, \text{head}_h)W^O
\end{equation}
where 
\begin{equation}
\begin{split}
\operatorname{head}_i=\operatorname{Attention}\left(Q W_Q^i, K W_K^i, V W_V^i\right), \\
\operatorname{Attention}(Q, K, V)=\sigma\left(\frac{Q K^{\top}+M}{\sqrt{d_{\text {head }}}}\right) V 
\end{split}
\end{equation}
where DropPath represents the stochastic depth regularization technique \cite{FractalNet}, and LayerNorm denotes the layer normalization method \cite{layernormalization}, which provides batch size-independent normalization capabilities. \textit{2) Feed-forward network:}
\begin{equation}
    O=\operatorname{DropPath}(\operatorname{MLP}(\operatorname{LayerNorm}(O))) \oplus X_{\text {out }},
\end{equation}
 \begin{equation}
F=\operatorname{Lin}\left(\operatorname{GELU}\left(\operatorname{DWConv} 2 \mathrm{D}_{3 \times 3}(\operatorname{GELU}(\mathrm{O}))\right)\right)+\mathrm{O}
 \end{equation}
where $\mathrm{DWConv2D}_{3 \times 3}$ denotes the \textit{3×3} depthwise convolution operation that introduces spatial locality inductive bias while maintaining computational efficiency through minimal parameter overhead. The SAM2-tiny backbone architecture comprises four Hierarchical stages containing $[1, 2, 7, 2]$ Hiera blocks respectively. Global attention mechanisms are implemented at blocks $[5, 7, 9, 11]$ to capture long-range spatial dependencies. \\
\textbf{Attention Window:} In the early stages, Hiera employs local windowed attention with a masking strategy where $M_{ij}=0$ for tokens $i$ and $j$ within the same mask unit, and $M_{ij}=-\infty$ otherwise. Later stages utilize global attention with $M=0$. Given the range-view image resolution of 64 × 2048, we propose an asymmetric attention window design that emphasizes horizontal spatial relationships: $8 \times64$ for the first and fourth stages, and $16\times128$ for the second and third stages. This horizontally-elongated window configuration, while empirically motivated, demonstrates superior performance compared to conventional square attention windows, effectively capturing the inherent horizontal structure of range-projected LiDAR data. \\
\textbf{Decoder:}
While the outputs of the encoder are multiscale features $F_1,F_2,F_3,F_4$ with spatial size,
\begin{equation}
    \left[(H, W),\left(\frac{H}{2}, \frac{W}{2}\right),\left(\frac{H}{4}, \frac{W}{4}\right),\left(\frac{H}{8}, \frac{W}{8}\right)\right]
\end{equation}

\begin{table*}[!htbp]
\centering
\setlength{\tabcolsep}{3pt}
\scriptsize
\caption{Comparisons among different Hiera backbones on the test set of SemanticKITTI \cite{semKitti1}. Interestingly, the SAM2-tiny model outperforms its larger counterparts despite having fewer parameters, demonstrating that model capacity does not necessarily correlate with performance in this task domain.}
\label{tab:models}
\resizebox{0.8\textwidth}{!}{%
\begin{tabular}{llrrrrrrrrrrrrrrrrrrrr}
\toprule
Backbone & Params(M) &  mIoU & car & bicy & moto & truc & o.veh & ped & b.list & m.list & road & park & walk & o.gro & build & fenc & veg & trun & terr & pole & sign \\
\midrule
\multicolumn{21}{l}{}\\

Hiera-tiny & 63.3 &\cellcolor{green!40}\textbf{61.5} & \cellcolor{green!40}95.0 &44.3 &47.3 &60.0 &\cellcolor{green!40}55.2& \cellcolor{green!40}68.2 &\cellcolor{green!40}79.6& \cellcolor{green!40}2.6 &94.8 &\cellcolor{green!40}49.6 &\cellcolor{green!40}82.1 &\cellcolor{green!40}1.2 &\cellcolor{green!40}88.6& \cellcolor{green!40}62.0 &85.1 &68.2 &71.3 &\cellcolor{green!40}64.8 &\cellcolor{green!40}48.2 \\
Hiera-small& 70.2 &60.5 &94.7 &\cellcolor{green!40}47.2 &\cellcolor{green!40}47.7 &\cellcolor{green!40}72.2 &39.7&63.1 &82.5 &0.0 &\cellcolor{green!40}95.3 &32.1& 80.8 &0.1 &88.3 &60.0& \cellcolor{green!40}88.4& \cellcolor{green!40}68.6 &\cellcolor{green!40}78.4 &64.9 &45.5\\

\bottomrule
\end{tabular}%
}
\vspace{-0.5em}
\end{table*}

\begin{table*}[!htbp]
\centering
\setlength{\tabcolsep}{3pt}
\scriptsize
\caption{Experiments with and without training augmentations. We observe a strong performance gain when introducing the proposed training augmentations from \cite{RangeFormer}. Comparisons among SAM2-tiny on SemanticKITTI test set \cite{semKitti1}.}
\label{tab:range_augs}
\resizebox{0.8\textwidth}{!}{%
\begin{tabular}{lrrrrrrrrrrrrrrrrrrrr}
\toprule
Augs. &  mIoU & car & bicy & moto & truc & o.veh & ped & b.list & m.list & road & park & walk & o.gro & build & fenc & veg & trun & terr & pole & sign \\
\midrule
\multicolumn{21}{l}{}\\

off  &55.0 &93.1 &36.2 &37.2& 53.9 &36.7 &53.7 &59.7 &0.0 &93.4 &35.3 &79.3 &0.0 &85.5 &55.8 &86.4 &59.3 &\cellcolor{green!40}76.7 &57.1& 45.3 \\
on &\cellcolor{green!40}\textbf{61.5}& \cellcolor{green!40}95.0 &\cellcolor{green!40}44.3 &\cellcolor{green!40}47.3 &\cellcolor{green!40}60.0& \cellcolor{green!40}55.2 &\cellcolor{green!40}68.2& \cellcolor{green!40}79.6 &\cellcolor{green!40}2.6 &\cellcolor{green!40}94.8 &\cellcolor{green!40}49.6 &\cellcolor{green!40}82.1 &\cellcolor{green!40}1.2 &\cellcolor{green!40}88.6 &\cellcolor{green!40}62.0 &\cellcolor{green!40}85.1 &\cellcolor{green!40}68.2 &71.3 &\cellcolor{green!40}64.8 &\cellcolor{green!40}48.2 \\

\bottomrule
\end{tabular}%
}
\vspace{-0.5em}
\end{table*}
\begin{table*}[!htbp]
\centering
\setlength{\tabcolsep}{3pt}
\scriptsize
\caption{Experiments with and without transfer-learning techniques using Cityscapes dataset \cite{Cityscapes}. We observe that overall our model does not benefit from extensive transfer-learning strategies introduced by \cite{RangeViT}. Comparisons among SAM2-tiny on SemanticKITTI \cite{semKitti1} test set.}
\label{tab:pretrained}
\resizebox{0.8\textwidth}{!}{%
\begin{tabular}{lrrrrrrrrrrrrrrrrrrrr}
\toprule
Pretrain &  mIoU & car & bicy & moto & truc & o.veh & ped & b.list & m.list & road & park & walk & o.gro & build & fenc & veg & trun & terr & pole & sign \\
\midrule
\multicolumn{21}{l}{}\\

yes &59.7 &94.2& 43.2& 38.0& \cellcolor{green!40}82.2 &45.3& 56.4 &65.5& 0.0 &\cellcolor{green!40}94.8 &43.7 &81.9& 0.1& 87.9 &60.7 &86.0 &68.10& \cellcolor{green!40}73.7& 64.3& 46.0 \\
no &\cellcolor{green!40}\textbf{61.5}& \cellcolor{green!40}95.0 &\cellcolor{green!40}44.3 &\cellcolor{green!40}47.3 &60.0 &\cellcolor{green!40}55.2 &\cellcolor{green!40}68.2 &\cellcolor{green!40}79.6& \cellcolor{green!40}2.6& \cellcolor{green!40}94.8 &\cellcolor{green!40}49.6& \cellcolor{green!40}82.1& \cellcolor{green!40}1.2 &\cellcolor{green!40}88.6 &\cellcolor{green!40}62.0 &8\cellcolor{green!40}5.1 &\cellcolor{green!40}68.2 &71.3 &\cellcolor{green!40}64.8& \cellcolor{green!40}48.2 \\

\bottomrule
\end{tabular}%
}
\vspace{-0.5em}
\end{table*}

\begin{table*}[!htbp]
\centering
\setlength{\tabcolsep}{3pt}
\scriptsize
\caption{Comparisons among state-of-the-art LiDAR range view semantic segmentation approaches on the validation sequence of SemanticKITTI \cite{semKitti1}. We compare our best performing model using SAM2 with the tiny Hiera backbone. Notably, our model is the first and only approach that leverages VFMs for this task.}
\label{tab:semkitti_range_table2}
\resizebox{0.9\textwidth}{!}{%
\begin{tabular}{lrrrrrrrrrrrrrrrrrrrrr}
\toprule
Method & VFM &  mIoU & car & bicy & moto & truc & o.veh & ped & b.list & m.list & road & park & walk & o.gro & build & fenc & veg & trun & terr & pole & sign \\
\midrule
\multicolumn{21}{l}{}\\

KPRNet\cite{kochanov2020kprnet} (2020)& $\times$ & 63.1 & \cellcolor{green!20}95.5 & 54.1 & 47.9 & 23.6 & 42.6 & 65.9 & 65.0 & 16.5 & \cellcolor{green!40}93.2 & \cellcolor{green!40}73.9 & \cellcolor{green!40}80.6 & 30.2 & 91.7 & 68.4 & \cellcolor{green!20}85.7 & 69.8 & \cellcolor{green!20}71.2 & 58.7 & 64.1 \\
LiteHDSeg\cite{razani2021litehdseg} (ICRA21) & $\times$ & 63.8 & 92.3 & 40.0 & 55.4 & 37.7 & 39.6 & 59.2 & \cellcolor{green!20}71.6 & \cellcolor{green!20}54.3 & \cellcolor{green!20}93.0 & 68.2 & 78.3 & 29.3 & 91.5 & 65.0 & 78.2 & 65.8 & 65.1 & 59.5 & \cellcolor{green!40}67.7 \\
MPF\cite{alnaggar2020multi} (WACV21)& $\times$ & 55.5 & 93.4 & 30.2 & 38.3 & 26.1 & 28.5 & 48.1 & 46.1 & 18.1 & 90.6 & 62.3 & 74.5 & 30.6 & 88.5 & 59.7 & 83.5 & 59.7 & 69.2 & 49.7 & 58.1 \\
FIDNet\cite{zhao2021fidnet} (IROS21) & $\times$ & 59.5 & 93.9 & 54.7 & 48.9 & 27.6 & 23.9 & 62.3 & 59.8 & 23.7 & 90.6 & 59.1 & 75.8 & 26.7 & 88.9 & 60.5 & 84.5 & 64.4 & 69.0 & 53.3 & 62.8 \\
RangeViT\cite{xu2021rangevit} (ICCV21) & $\times$ & 64.0 & 95.4 & 55.8 & 43.5 & 29.8 & 42.1 & 63.9 & 58.2 & 38.1 & 93.1 & 70.2 & \cellcolor{green!20}80.0 & \cellcolor{green!20}32.5 & \cellcolor{green!20}92.0 & \cellcolor{green!20}69.0 & 85.3 & \cellcolor{green!20}70.6 & \cellcolor{green!20}71.2 & 60.8 & 64.7 \\
CENet\cite{cheng2022cenet} (ICME22) & $\times$ & 64.7 & 91.9 & \cellcolor{green!20}58.6 & 50.3 & 40.6 & 42.3 & \cellcolor{green!20}68.9 & 65.9 & 43.5 & 90.3 & 60.9 & 75.1 & 31.5 & 91.0 & 66.2 & 84.5 & 69.7 & 70.0 & \cellcolor{green!20}61.5 & 67.6 \\
MaskRange\cite{Gu2022MaskRange} (2022) & $\times$ & \cellcolor{green!20}66.1 & 94.2 & 56.0 & \cellcolor{green!20}55.7 & \cellcolor{green!20}59.2 & \cellcolor{green!20}52.4 & 67.6 & 64.8 & 31.8 & 91.7 & 70.7 & 77.1 & 29.5 & 90.6 & 65.2 & 84.6 & 68.5 & 69.2 & 60.2 & \cellcolor{green!20}66.6 \\
RangeFormer\cite{zhang2022rangeformer} (NeurIPS22) & $\times$  & \cellcolor{green!40}73.3 & \cellcolor{green!40}96.7 & \cellcolor{green!40}69.4 & \cellcolor{green!40}73.7 & \cellcolor{green!40}59.9 & \cellcolor{green!40}66.2 & \cellcolor{green!40}78.1 & \cellcolor{green!40}75.9 & \cellcolor{green!40}58.1 & 92.4 & \cellcolor{green!20}73.0 & 78.8 & \cellcolor{green!40}42.4 & \cellcolor{green!40}92.3 & \cellcolor{green!40}70.1 & \cellcolor{green!40}86.6 & \cellcolor{green!40}73.3 & \cellcolor{green!40}72.8 & \cellcolor{green!40}66.4 & \cellcolor{green!20}66.6 \\
\textbf{RangeSAM(Ours)} & \checkmark & \textbf{60.9} & \cellcolor{blue!20}91.7 & 42.0 & 44.3 & 41.2 & 36.1 & 55.9 & 54.4 & 30.2 & \cellcolor{blue!20}92.2 & \cellcolor{blue!20}67.5 & \cellcolor{blue!20}77.2 & \cellcolor{blue!20}28.6 & \cellcolor{blue!20}89.8 & 62.8 & \cellcolor{blue!20}84.5 & \cellcolor{blue!20}66.9 & \cellcolor{blue!20}69.8 & \cellcolor{blue!20}58.9 & \cellcolor{blue!20}62.4 \\
\bottomrule
\end{tabular}%
}
\vspace{-0.5em}
\end{table*}
and channel dimensions $[96, 192, 384, 768]$. Following \cite{sam2unet}, we standardized the output channel dimensions to 256 using a modified Receptive Field Block \cite{ReceptiveFieldBlock} architecture that substitutes LayerNorm and GELU activations for the conventional BatchNorm \cite{BatchNorm} and ReLU, thereby achieving better compatibility with contemporary transformer architectures. Consistent with \cite{KITTIVision}, we concatenate these four normalized feature maps and progressively reduce the dimensionality to match the number of target classes while incorporating auxiliary classification heads at corresponding feature levels to enhance gradient flow during training.\\
\textbf{Postprocessing:}
Dense evaluation on datasets like SemanticKITTI requires label propagation from processed points to the full-resolution point cloud via \emph{k-NN} interpolation with majority voting. We use $k = 7$, balancing efficiency and noise robustness within the typical range of 3-7 \cite{RangeSeg, MetaRangeSeg, RangeViT, RangeFormer}.\\
\textbf{Loss:}
We follow the training objective of the proposed SAM2-UNet \cite{sam2unet} model which incorporates a multi-component loss function comprising weighted cross-entropy ($\mathcal{L}_{W C E}$), Dice loss ($\mathcal{L}_{\text {Dice }}$), boundary loss ($\mathcal{L}_{\text {Boundary }}$) and Jaccard index loss ($\mathcal{L}_{\text {IoU }}$). This composite loss formulation addresses class imbalance through weighted cross-entropy \cite{WCELoss}, enhances region-level segmentation accuracy via Dice and Jaccard losses \cite{diceLoss, wang2024jaccardmetriclossesoptimizing}, and promotes precise boundary delineation through the boundary loss \cite{boundaryLoss} component.
The combined loss function is expressed as a weighted sum:
\begin{equation}
    \mathcal{L}_{\text {total }}=\lambda_1 \mathcal{L}_{W C E}+\lambda_2 \mathcal{L}_{\text {Dice }}+\lambda_3 \mathcal{L}_{\text {Boundary }}+\lambda_4 \mathcal{L}_{\text {IoU }},
\end{equation}
where $\lambda_i$ represent empirically determined weighting coefficients. In our experimental configuration, we set $\lambda_i = 1$, assigning equal importance to each loss component.

\section{Dataset and implementation details}
\label{sec:exps}
We now describe the datasets used, the training configurations applied, and the evaluation metrics employed.
\subsection{Dataset selection and metrics:}
We used SemanticKITTI \cite{semKitti1} and nuScenes \cite{nuScenes} as our primary training datasets. SemanticKITTI consisted of urban driving scenes captured with a 64-beam LiDAR, split into 19,130 training scans (sequences 00-07, 09-10), 4,071 validation scans (sequence 08), and 20,351 test scans (sequences 11-21) across 19 semantic classes. NuScenes contained 1,000 urban scenes from Boston and Singapore captured with a 32-beam LiDAR, providing 28,130 training and 6,019 validation frames across 16 categories. We converted nuScenes to SemanticKITTI format for consistency and reported mean Intersection over Union (mIoU) \cite{miou} on the SemanticKITTI validation split.
\subsection{Data Augmentations:}
We applied standard augmentations including global rotation, coordinate jittering, flipping, and random point elimination (all with probability 1.0), along with range-view-specific augmentations from \cite{RangeFormer}, mixing, union, shifting, and copy-paste with probabilities [0.9, 0.1, 0.9, 1.0], respectively. 
\subsection{Training strategy:}
All experiments used 8 A100 GPUs (40GB) with a per-GPU batch size of 2 (effective batch size of 16). Training employed a 5-epoch linear warm-up followed by cosine annealing \cite{ACloserLook}. Following recent works \cite{RangeFormer, RangeViT, ptv3, ppt} show that transformer-based architectures benefit from pretraining, we adopted the training protocol from \cite{RangeFormer, RangeViT} and pretrain on nuScenes \cite{nuScenes} (converted to SemanticKITTI format) for 60 epochs. We selected SAM2-tiny as our backbone, as heavier SAM2 variants yielded negligible performance gains (Table \ref{tab:models}) while significantly increasing computational cost and inference latency. The final model contains approximately 63 million parameters.
\textbf{Optimization:}
We use AdamW with differentiated learning rates: the Hiera backbone uses $lr=0.0004$ and weight decay $\alpha=0.001$, while the rest of the model uses $lr=0.001$ and $\alpha=0.0001$.

\subsection{Evaluation}
We present RangeSAM's experimental results on SemanticKITTI \cite{semKitti1}. Table \ref{tab:semkitti_range_table2} shows validation results on sequence 08, while Tables \ref{tab:models}, \ref{tab:range_augs}, and \ref{tab:pretrained} report test set performance. Our approach achieves competitive results compared to state-of-the-art methods.

\begin{figure}[ht]
 \centering\includegraphics[width=\linewidth]{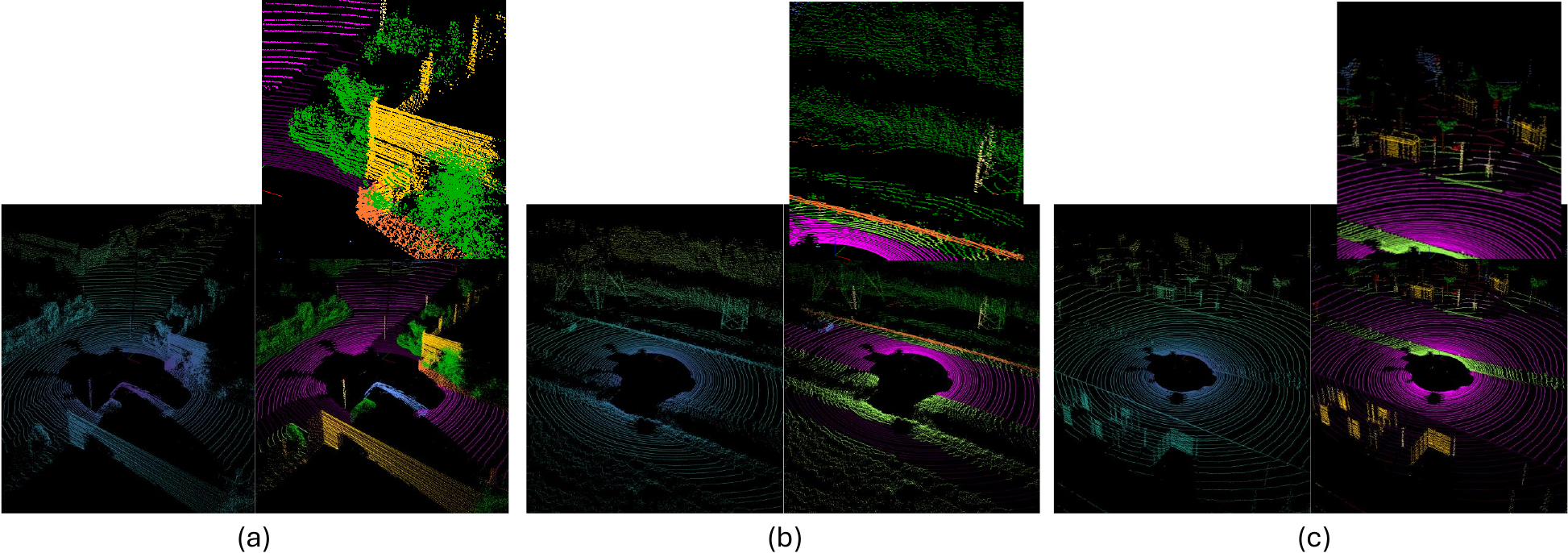}
  \caption{Qualitative segmentation examples of increasing difficulty using our model. (a) Urban intersection, (b) Suburban environment and (c) Highly cluttered scene.}
  \label{fig:qualitative_analysis}
\end{figure}
\subsection{Quantitative evaluation}
Table \ref{tab:models} compares SAM2 Hiera backbones: tiny and small variants show comparable mIoU, with tiny offering computational advantages through reduced parameters and faster inference.

Table \ref{tab:semkitti_range_table2} benchmarks our model against baselines. While recent state-of-the-art achieves $65–75\%$ mIoU, our model reaches $60.9\%$, successfully integrating the SAM2 VFM~\cite{SAM2}. Performance stratifies across three tiers: \begin{itemize} \item High IoU ($80–90\%$): Large, frequent classes (cars, road, building, vegetation) perform competitively with state-of-the-art. \item Mid IoU ($60–70\%$): Mid-frequency categories (trucks, fences, terrain) remain competitive. \item Low IoU ($29–47\%$): Rare, small objects (motorcycles, bicycles, persons) prove challenging, consistent with modern approaches. \end{itemize}
While lower performance on long-tail categories reflects limited training examples relative to our 63M parameters, results remain competitive on well-represented categories (\colorbox{blue!20}{highlighted} in Table \ref{tab:semkitti_range_table2}). Dataset expansion and design refinement should close the gap on tail classes.

\subsection{Qualitative evaluation}
Figure~\ref{fig:qualitative_analysis} illustrates three segmentation scenarios: \textbf{(a) Urban intersection:} The model accurately segments vehicles, static elements (trees, walls, fences), and fine structures (fence posts). \textbf{(b) Suburban environment:} Dominant classes (cars, vegetation, fences) are correctly segmented, but the model struggles with unseen classes like transmission towers. \textbf{(c) Highly cluttered scene:} Despite overlapping objects (trees, fences, poles, signs), major classes are captured. Small or distant structures exhibit noise, but overall segmentation remains robust.

\subsection{Ablation Study}
We conduct a comprehensive ablation study to systematically evaluate the impact of different training strategies on model performance.
\\
\textbf{Transfer learning}: We followed a transfer learning approach established by other baseline methods\cite{RangeViT} in this area of research. For the experiments in Table \ref{tab:pretrained}, we pretrained our model on the Cityscapes \cite{Cityscapes} dataset for 25 epochs to enhance feature learning before training on the target dataset. However, this approach led to reduced performance. We believe this may be due to the SAM2 model being extensively pretrained on large-scale image datasets, which could have caused a domain mismatch affecting transfer learning. Further investigation is needed to confirm this hypothesis and explore other adaptation strategies.\\
\textbf{Range View Augmentations:} The results (see Table~\ref{tab:range_augs}) indicate that adding augmentations introduced by \cite{RangeFormer} to the training process improves performance across nearly all datasets, achieving a 10\% gain in mIoU.

\section{Conclusion}
\label{sec:conclusion}
This work, to our knowledge, is the first to adapt VFMs for range-view LiDAR point cloud segmentation. Despite the domain gap between RGB images and range-view representations, our targeted architectural modifications to the SAM2 encoder and decoder effectively bridged this modality shift and achieved competitive performance.
We have integrated augmentation strategies and training protocols from prior work \cite{RangeSeg, RangeFormer, ptv3, RangeViT}. Notably, we find that multi-dataset training across diverse 3D benchmarks outperforms extensive 2D pretraining as in \cite{RangeViT}. We will release source code and model weights.\\
\textbf{Future Work:} The primary limitation is computational complexity. The convolution-free Hiera backbone \cite{Hiera} struggles with point cloud sparsity, requiring RFBs \cite{ReceptiveFieldBlock} for competitive performance. However, RFBs constitute the main computational bottleneck, currently preventing real-time deployment—a key target for future optimization.



{
    \small
    \bibliographystyle{ieeenat_fullname}
    \bibliography{main}

\begin{thebibliography}{98}
\providecommand{\natexlab}[1]{#1}
\providecommand{\url}[1]{\texttt{#1}}
\expandafter\ifx\csname urlstyle\endcsname\relax
  \providecommand{\doi}[1]{doi: #1}\else
  \providecommand{\doi}{doi: \begingroup \urlstyle{rm}\Url}\fi

\bibitem[Alnaggar et~al.(2021)Alnaggar, Afifi, Amer, and Elhelw]{alnaggar2020multi}
Yara~Ali Alnaggar, Mohamed Afifi, Karim Amer, and Mohamed Elhelw.
\newblock {Multi-Projection Fusion for Real-Time Semantic Segmentation of 3D LiDAR Point Clouds}.
\newblock In \emph{Winter Conference on Applications of Computer Vision (WACV)}, 2021.

\bibitem[Ando et~al.(2023)Ando, Gidaris, Bursuc, Puy, Boulch, and Marlet]{RangeViT}
Angelika Ando, Spyros Gidaris, Andrei Bursuc, Gilles Puy, Alexandre Boulch, and Renaud Marlet.
\newblock Rangevit: Towards vision transformers for 3d semantic segmentation in autonomous driving.
\newblock In \emph{2023 IEEE/CVF Conference on Computer Vision and Pattern Recognition (CVPR)}, pages 5240--5250, 2023.

\bibitem[Ba et~al.(2016)Ba, Kiros, and Hinton]{layernormalization}
Jimmy~Lei Ba, Jamie~Ryan Kiros, and Geoffrey~E. Hinton.
\newblock Layer normalization, 2016.

\bibitem[Behley et~al.(2019)Behley, Garbade, Milioto, Quenzel, Behnke, Stachniss, and Gall]{semKitti1}
J. Behley, M. Garbade, A. Milioto, J. Quenzel, S. Behnke, C. Stachniss, and J. Gall.
\newblock {SemanticKITTI: A Dataset for Semantic Scene Understanding of LiDAR Sequences}.
\newblock In \emph{Proc. of the IEEE/CVF International Conf.~on Computer Vision (ICCV)}, 2019.

\bibitem[Behley et~al.(2021)Behley, Garbade, Milioto, Quenzel, Behnke, Gall, and Stachniss]{semKitti2}
J. Behley, M. Garbade, A. Milioto, J. Quenzel, S. Behnke, J. Gall, and C. Stachniss.
\newblock {Towards 3D LiDAR-based semantic scene understanding of 3D point cloud sequences: The SemanticKITTI Dataset}.
\newblock \emph{The International Journal on Robotics Research}, 40\penalty0 (8-9):\penalty0 959--967, 2021.

\bibitem[Bommasani et~al.(2021)Bommasani, Hudson, Adeli, Altman, Arora, von Arx, Bernstein, Bohg, Bosselut, Brunskill, Brynjolfsson, Buch, Card, Castellon, Chatterji, Chen, Creel, Davis, Demszky, Donahue, Doumbouya, Durmus, Ermon, Etchemendy, Ethayarajh, Fei{-}Fei, Finn, Gale, Gillespie, Goel, Goodman, Grossman, Guha, Hashimoto, Henderson, Hewitt, Ho, Hong, Hsu, Huang, Icard, Jain, Jurafsky, Kalluri, Karamcheti, Keeling, Khani, Khattab, Koh, Krass, Krishna, Kuditipudi, and et~al.]{OnTheOppsOfVFMs}
Rishi Bommasani, Drew~A. Hudson, Ehsan Adeli, Russ~B. Altman, Simran Arora, Sydney von Arx, Michael~S. Bernstein, Jeannette Bohg, Antoine Bosselut, Emma Brunskill, Erik Brynjolfsson, Shyamal Buch, Dallas Card, Rodrigo Castellon, Niladri~S. Chatterji, Annie~S. Chen, Kathleen Creel, Jared~Quincy Davis, Dorottya Demszky, Chris Donahue, Moussa Doumbouya, Esin Durmus, Stefano Ermon, John Etchemendy, Kawin Ethayarajh, Li Fei{-}Fei, Chelsea Finn, Trevor Gale, Lauren~E. Gillespie, Karan Goel, Noah~D. Goodman, Shelby Grossman, Neel Guha, Tatsunori Hashimoto, Peter Henderson, John Hewitt, Daniel~E. Ho, Jenny Hong, Kyle Hsu, Jing Huang, Thomas Icard, Saahil Jain, Dan Jurafsky, Pratyusha Kalluri, Siddharth Karamcheti, Geoff Keeling, Fereshte Khani, Omar Khattab, Pang~Wei Koh, Mark~S. Krass, Ranjay Krishna, Rohith Kuditipudi, and et al.
\newblock On the opportunities and risks of foundation models.
\newblock \emph{CoRR}, abs/2108.07258, 2021.

\bibitem[Bui et~al.(2024)Bui, Hoang, Tran, Doretto, Adjeroh, Patel, Choudhary, and Le]{sam3dsegmentMedical}
Nhat-Tan Bui, Dinh-Hieu Hoang, Minh-Triet Tran, Gianfranco Doretto, Donald Adjeroh, Brijesh Patel, Arabinda Choudhary, and Ngan Le.
\newblock Sam3d: Segment anything model in volumetric medical images, 2024.

\bibitem[Caesar et~al.(2019)Caesar, Bankiti, Lang, Vora, Liong, Xu, Krishnan, Pan, Baldan, and Beijbom]{nuScenes}
Holger Caesar, Varun Bankiti, Alex~H. Lang, Sourabh Vora, Venice~Erin Liong, Qiang Xu, Anush Krishnan, Yu Pan, Giancarlo Baldan, and Oscar Beijbom.
\newblock nuscenes: {A} multimodal dataset for autonomous driving.
\newblock \emph{CoRR}, abs/1903.11027, 2019.

\bibitem[Caron et~al.(2021)Caron, Touvron, Misra, J{\'{e}}gou, Mairal, Bojanowski, and Joulin]{DINOv1}
Mathilde Caron, Hugo Touvron, Ishan Misra, Herv{\'{e}} J{\'{e}}gou, Julien Mairal, Piotr Bojanowski, and Armand Joulin.
\newblock Emerging properties in self-supervised vision transformers.
\newblock \emph{CoRR}, abs/2104.14294, 2021.

\bibitem[Chen et~al.(2024{\natexlab{a}})Chen, Xia, Mao, Wang, and Zhang]{sieseffusionnetspatialintercorrelationenhancement}
Jiale Chen, Fei Xia, Jianliang Mao, Haoping Wang, and Chuanlin Zhang.
\newblock Siesef-fusionnet: Spatial inter-correlation enhancement and spatially-embedded feature fusion network for lidar point cloud semantic segmentation, 2024{\natexlab{a}}.

\bibitem[Chen et~al.(2023)Chen, Liu, Kong, Zhu, Ma, Li, Hou, Qiao, and Wang]{clip2scene}
Runnan Chen, Youquan Liu, Lingdong Kong, Xinge Zhu, Yuexin Ma, Yikang Li, Yuenan Hou, Yu Qiao, and Wenping Wang.
\newblock Clip2scene: Towards label-efficient 3d scene understanding by clip, 2023.

\bibitem[Chen et~al.(2024{\natexlab{b}})Chen, Lu, Zhu, Ding, Yu, Ji, Li, Sun, Mao, and Zang]{SAM2Adapter}
Tianrun Chen, Ankang Lu, Lanyun Zhu, Chaotao Ding, Chunan Yu, Deyi Ji, Zejian Li, Lingyun Sun, Papa Mao, and Ying Zang.
\newblock Sam2-adapter: Evaluating \& adapting segment anything 2 in downstream tasks: Camouflage, shadow, medical image segmentation, and more, 2024{\natexlab{b}}.

\bibitem[Chen and Chang(2022)]{RangeSeg}
Tzu-Hsuan Chen and Tian~Sheuan Chang.
\newblock Rangeseg: Range-aware real time segmentation of 3d lidar point clouds.
\newblock \emph{IEEE Transactions on Intelligent Vehicles}, 7\penalty0 (1):\penalty0 93–101, 2022.

\bibitem[Cheng et~al.(2022)Cheng, Han, and Xiao]{cheng2022cenet}
Hui-Xian Cheng, Xian-Feng Han, and Guo-Qiang Xiao.
\newblock Cenet: Toward concise and efficient lidar semantic segmentation for autonomous driving.
\newblock In \emph{2022 IEEE International Conference on Multimedia and Expo (ICME)}, pages 01--06. IEEE, 2022.

\bibitem[Cordts et~al.(2016)Cordts, Omran, Ramos, Rehfeld, Enzweiler, Benenson, Franke, Roth, and Schiele]{Cityscapes}
Marius Cordts, Mohamed Omran, Sebastian Ramos, Timo Rehfeld, Markus Enzweiler, Rodrigo Benenson, Uwe Franke, Stefan Roth, and Bernt Schiele.
\newblock The cityscapes dataset for semantic urban scene understanding.
\newblock In \emph{Proc. of the IEEE Conference on Computer Vision and Pattern Recognition (CVPR)}, 2016.

\bibitem[Cortinhal et~al.(2020)Cortinhal, Tzelepis, and Aksoy]{cortinhal2020salsanext}
Tiago Cortinhal, George Tzelepis, and Eren~Erdal Aksoy.
\newblock Salsanext: Fast, uncertainty-aware semantic segmentation of lidar point clouds for autonomous driving, 2020.

\bibitem[Dhawan et~al.(2019)Dhawan, Bodani, and Garg]{CRF}
Aashish Dhawan, Pankaj Bodani, and Vishal Garg.
\newblock Post processing of image segmentation using conditional random fields.
\newblock In \emph{2019 6th International Conference on Computing for Sustainable Global Development (INDIACom)}, pages 729--734, 2019.

\bibitem[Engelmann et~al.(2020)Engelmann, Bokeloh, Fathi, Leibe, and Nie{\ss}ner]{3D-MPA}
Francis Engelmann, Martin Bokeloh, Alireza Fathi, Bastian Leibe, and Matthias Nie{\ss}ner.
\newblock 3d-mpa: Multi proposal aggregation for 3d semantic instance segmentation.
\newblock \emph{CoRR}, abs/2003.13867, 2020.

\bibitem[Everingham et~al.(2015)Everingham, Eslami, Gool, Williams, Winn, and Zisserman]{miou}
Mark Everingham, S.~M. Eslami, Luc Gool, Christopher~K. Williams, John Winn, and Andrew Zisserman.
\newblock The pascal visual object classes challenge: A retrospective.
\newblock \emph{Int. J. Comput. Vision}, 111\penalty0 (1):\penalty0 98–136, 2015.

\bibitem[Fang et~al.(2022)Fang, Wang, Xie, Sun, Wu, Wang, Huang, Wang, and Cao]{EVA01}
Yuxin Fang, Wen Wang, Binhui Xie, Quan Sun, Ledell Wu, Xinggang Wang, Tiejun Huang, Xinlong Wang, and Yue Cao.
\newblock Eva: Exploring the limits of masked visual representation learning at scale.
\newblock \emph{arXiv preprint arXiv:2211.07636}, 2022.

\bibitem[Fang et~al.(2024)Fang, Sun, Wang, Huang, Wang, and Cao]{EVA02}
Yuxin Fang, Quan Sun, Xinggang Wang, Tiejun Huang, Xinlong Wang, and Yue Cao.
\newblock Eva-02: A visual representation for neon genesis.
\newblock \emph{Image and Vision Computing}, 149:\penalty0 105171, 2024.

\bibitem[Geiger et~al.(2012)Geiger, Lenz, and Urtasun]{KITTIVision}
Andreas Geiger, Philip Lenz, and Raquel Urtasun.
\newblock Are we ready for autonomous driving? the kitti vision benchmark suite.
\newblock In \emph{2012 IEEE Conference on Computer Vision and Pattern Recognition}, pages 3354--3361, 2012.

\bibitem[Gotmare et~al.(2018)Gotmare, Keskar, Xiong, and Socher]{ACloserLook}
Akhilesh Gotmare, Nitish~Shirish Keskar, Caiming Xiong, and Richard Socher.
\newblock A closer look at deep learning heuristics: Learning rate restarts, warmup and distillation.
\newblock \emph{CoRR}, abs/1810.13243, 2018.

\bibitem[Gu et~al.(2022)Gu, Huang, Xu, and Kong]{Gu2022MaskRange}
Yi Gu, Yuming Huang, Chengzhong Xu, and Hui Kong.
\newblock Maskrange: A mask-classification model for range-view based lidar segmentation, 2022.
\newblock Under review.

\bibitem[Hendrycks and Gimpel(2016)]{GELU}
Dan Hendrycks and Kevin Gimpel.
\newblock Bridging nonlinearities and stochastic regularizers with gaussian error linear units.
\newblock \emph{CoRR}, abs/1606.08415, 2016.

\bibitem[Hou et~al.(2022)Hou, Zhu, Ma, Loy, and Li]{PVKD}
Yuenan Hou, Xinge Zhu, Yuexin Ma, Chen~Change Loy, and Yikang Li.
\newblock Point-to-voxel knowledge distillation for lidar semantic segmentation, 2022.

\bibitem[Hu et~al.(2020)Hu, Yang, Xie, Rosa, Guo, Wang, Trigoni, and Markham]{RandLA}
Qingyong Hu, Bo Yang, Linhai Xie, Stefano Rosa, Yulan Guo, Zhihua Wang, Niki Trigoni, and Andrew Markham.
\newblock { RandLA-Net: Efficient Semantic Segmentation of Large-Scale Point Clouds }.
\newblock In \emph{2020 IEEE/CVF Conference on Computer Vision and Pattern Recognition (CVPR)}, pages 11105--11114, Los Alamitos, CA, USA, 2020. IEEE Computer Society.

\bibitem[Huo et~al.(2025)Huo, Dai, and Tang]{huo2025sambaunetsynergizingsam2mamba}
Guohao Huo, Ruiting Dai, and Hao Tang.
\newblock Samba-unet: Synergizing sam2 and mamba in unet with heterogeneous aggregation for cardiac mri segmentation, 2025.

\bibitem[Ioffe and Szegedy(2015)]{BatchNorm}
Sergey Ioffe and Christian Szegedy.
\newblock Batch normalization: accelerating deep network training by reducing internal covariate shift.
\newblock In \emph{Proceedings of the 32nd International Conference on International Conference on Machine Learning - Volume 37}, page 448–456. JMLR.org, 2015.

\bibitem[Kirillov et~al.(2023)Kirillov, Mintun, Ravi, Mao, Rolland, Gustafson, Xiao, Whitehead, Berg, Lo, Dollár, and Girshick]{SAM1}
Alexander Kirillov, Eric Mintun, Nikhila Ravi, Hanzi Mao, Chloe Rolland, Laura Gustafson, Tete Xiao, Spencer Whitehead, Alexander~C. Berg, Wan-Yen Lo, Piotr Dollár, and Ross Girshick.
\newblock Segment anything, 2023.

\bibitem[Kochanov et~al.(2020{\natexlab{a}})Kochanov, Nejadasl, and Booij]{Kochanov2020KPRNetIP}
Deyvid Kochanov, F.~Karimi Nejadasl, and Olaf Booij.
\newblock Kprnet: Improving projection-based lidar semantic segmentation.
\newblock \emph{ArXiv}, abs/2007.12668, 2020{\natexlab{a}}.

\bibitem[Kochanov et~al.(2020{\natexlab{b}})Kochanov, Nejadasl, and Booij]{kochanov2020kprnet}
Deyvid Kochanov, Fatemeh~Karimi Nejadasl, and Olaf Booij.
\newblock {KPRNet: Improving projection-based LiDAR semantic segmentation}.
\newblock \emph{arXiv preprint arXiv:2007.12668}, 2020{\natexlab{b}}.

\bibitem[Kong et~al.(2023)Kong, Liu, Chen, Ma, Zhu, Li, Hou, Qiao, and Liu]{RangeFormer}
Lingdong Kong, Youquan Liu, Runnan Chen, Yuexin Ma, Xinge Zhu, Yikang Li, Yuenan Hou, Yu Qiao, and Ziwei Liu.
\newblock Rethinking range view representation for lidar segmentation, 2023.

\bibitem[Landrieu and Simonovsky(2017)]{SuperpointGraphs}
Lo{\"{\i}}c Landrieu and Martin Simonovsky.
\newblock Large-scale point cloud semantic segmentation with superpoint graphs.
\newblock \emph{CoRR}, abs/1711.09869, 2017.

\bibitem[Larsson et~al.(2016)Larsson, Maire, and Shakhnarovich]{FractalNet}
Gustav Larsson, Michael Maire, and Gregory Shakhnarovich.
\newblock Fractalnet: Ultra-deep neural networks without residuals.
\newblock \emph{CoRR}, abs/1605.07648, 2016.

\bibitem[Li et~al.(2023)Li, Zhang, Sun, Zou, Liu, Yang, Li, Zhang, and Gao]{semanticSAM}
Feng Li, Hao Zhang, Peize Sun, Xueyan Zou, Shilong Liu, Jianwei Yang, Chunyuan Li, Lei Zhang, and Jianfeng Gao.
\newblock Semantic-sam: Segment and recognize anything at any granularity.
\newblock \emph{arXiv preprint arXiv:2307.04767}, 2023.

\bibitem[Liu et~al.(2017)Liu, Huang, and Wang]{ReceptiveFieldBlock}
Songtao Liu, Di Huang, and Yunhong Wang.
\newblock Receptive field block net for accurate and fast object detection.
\newblock \emph{CoRR}, abs/1711.07767, 2017.

\bibitem[Liu et~al.(2023)Liu, Zeng, Ren, Li, Zhang, Yang, Li, Yang, Su, Zhu, et~al.]{groundingDino}
Shilong Liu, Zhaoyang Zeng, Tianhe Ren, Feng Li, Hao Zhang, Jie Yang, Chunyuan Li, Jianwei Yang, Hang Su, Jun Zhu, et~al.
\newblock Grounding dino: Marrying dino with grounded pre-training for open-set object detection.
\newblock \emph{arXiv preprint arXiv:2303.05499}, 2023.

\bibitem[Liu et~al.(2025)Liu, Chen, Wang, Han, Wu, Peng, Chen, Tian, and Carneiro]{liu2025auralsam2enablingsam2hear}
Yuyuan Liu, Yuanhong Chen, Chong Wang, Junlin Han, Junde Wu, Can Peng, Jingkun Chen, Yu Tian, and Gustavo Carneiro.
\newblock Auralsam2: Enabling sam2 hear through pyramid audio-visual feature prompting, 2025.

\bibitem[Messmer et~al.(2024)Messmer, Reich, and Abdeslam]{ContextAwareML}
Liane-Marina Messmer, Christoph Reich, and Djaffar~Ould Abdeslam.
\newblock Context-aware machine learning: A survey.
\newblock In \emph{Proceedings of the Future Technologies Conference (FTC) 2024, Volume 1}, pages 252--272, Cham, 2024. Springer Nature Switzerland.

\bibitem[Milioto et~al.(2019)Milioto, Vizzo, Behley, and Stachniss]{RangeNet++}
Andres Milioto, Ignacio Vizzo, Jens Behley, and Cyrill Stachniss.
\newblock Rangenet ++: Fast and accurate lidar semantic segmentation.
\newblock In \emph{2019 IEEE/RSJ International Conference on Intelligent Robots and Systems (IROS)}, pages 4213--4220, 2019.

\bibitem[Oquab et~al.(2023)Oquab, Darcet, Moutakanni, Vo, Szafraniec, Khalidov, Fernandez, Haziza, Massa, El-Nouby, Assran, Ballas, Galuba, Howes, Huang, Li, Misra, Rabbat, Sharma, Synnaeve, Xu, J{\'e}gou, Mairal, Labatut, Joulin, and Bojanowski]{DINOv2}
Maxime Oquab, Timoth{\'e}e Darcet, Th{\'e}o Moutakanni, Huy~Q. Vo, Marc Szafraniec, Vasil Khalidov, Pierre Fernandez, Daniel Haziza, Francisco Massa, Alaaeldin El-Nouby, Mahmoud Assran, Nicolas Ballas, Wojciech Galuba, Russ Howes, Po-Yao~(Bernie) Huang, Shang-Wen Li, Ishan Misra, Michael~G. Rabbat, Vasu Sharma, Gabriel Synnaeve, Huijiao Xu, Herv{\'e} J{\'e}gou, Julien Mairal, Patrick Labatut, Armand Joulin, and Piotr Bojanowski.
\newblock Dinov2: Learning robust visual features without supervision.
\newblock \emph{ArXiv}, abs/2304.07193, 2023.

\bibitem[Pan et~al.(2021)Pan, Xia, Song, Li, and Huang]{ptv1}
Xuran Pan, Zhuofan Xia, Shiji Song, Li~Erran Li, and Gao Huang.
\newblock 3d object detection with pointformer.
\newblock In \emph{2021 IEEE/CVF Conference on Computer Vision and Pattern Recognition (CVPR)}, pages 7459--7468, 2021.

\bibitem[Pan et~al.(2020)Pan, Gao, Mei, Geng, Li, and Zhao]{semanticPOSS}
Yancheng Pan, Biao Gao, Jilin Mei, Sibo Geng, Chengkun Li, and Huijing Zhao.
\newblock Semanticposs: {A} point cloud dataset with large quantity of dynamic instances.
\newblock \emph{CoRR}, abs/2002.09147, 2020.

\bibitem[Peng et~al.(2023)Peng, Genova, Jiang, Tagliasacchi, Pollefeys, and Funkhouser]{OpenScene}
Songyou Peng, Kyle Genova, Chiyu Jiang, Andrea Tagliasacchi, Marc Pollefeys, and Thomas Funkhouser.
\newblock Openscene: 3d scene understanding with open vocabularies.
\newblock In \emph{2023 IEEE/CVF Conference on Computer Vision and Pattern Recognition (CVPR)}, pages 815--824, 2023.

\bibitem[Pham et~al.(2019)Pham, Nguyen, Hua, Roig, and Yeung]{JSIS3D}
Quang{-}Hieu Pham, Duc~Thanh Nguyen, Binh{-}Son Hua, Gemma Roig, and Sai{-}Kit Yeung.
\newblock {JSIS3D:} joint semantic-instance segmentation of 3d point clouds with multi-task pointwise networks and multi-value conditional random fields.
\newblock \emph{CoRR}, abs/1904.00699, 2019.

\bibitem[Qi et~al.(2016)Qi, Su, Mo, and Guibas]{PointNet}
Charles~Ruizhongtai Qi, Hao Su, Kaichun Mo, and Leonidas~J. Guibas.
\newblock Pointnet: Deep learning on point sets for 3d classification and segmentation.
\newblock \emph{CoRR}, abs/1612.00593, 2016.

\bibitem[Qi et~al.(2017{\natexlab{a}})Qi, Liu, Wu, Su, and Guibas]{FrustumPointNets}
Charles~Ruizhongtai Qi, Wei Liu, Chenxia Wu, Hao Su, and Leonidas~J. Guibas.
\newblock Frustum pointnets for 3d object detection from {RGB-D} data.
\newblock \emph{CoRR}, abs/1711.08488, 2017{\natexlab{a}}.

\bibitem[Qi et~al.(2017{\natexlab{b}})Qi, Yi, Su, and Guibas]{PointNetPlusPlus}
Charles~Ruizhongtai Qi, Li Yi, Hao Su, and Leonidas~J. Guibas.
\newblock Pointnet++: Deep hierarchical feature learning on point sets in a metric space.
\newblock \emph{CoRR}, abs/1706.02413, 2017{\natexlab{b}}.

\bibitem[Ravi et~al.(2024)Ravi, Gabeur, Hu, Hu, Ryali, Ma, Khedr, Rädle, Rolland, Gustafson, Mintun, Pan, Alwala, Carion, Wu, Girshick, Dollár, and Feichtenhofer]{SAM2}
Nikhila Ravi, Valentin Gabeur, Yuan-Ting Hu, Ronghang Hu, Chaitanya Ryali, Tengyu Ma, Haitham Khedr, Roman Rädle, Chloe Rolland, Laura Gustafson, Eric Mintun, Junting Pan, Kalyan~Vasudev Alwala, Nicolas Carion, Chao-Yuan Wu, Ross Girshick, Piotr Dollár, and Christoph Feichtenhofer.
\newblock Sam 2: Segment anything in images and videos, 2024.

\bibitem[Razani et~al.(2021{\natexlab{a}})Razani, Cheng, Taghavi, and Bingbing]{LiteHDNet}
Ryan Razani, Ran Cheng, Ehsan Taghavi, and Liu Bingbing.
\newblock Lite-hdseg: Lidar semantic segmentation using lite harmonic dense convolutions.
\newblock In \emph{2021 IEEE International Conference on Robotics and Automation (ICRA)}, pages 9550--9556, 2021{\natexlab{a}}.

\bibitem[Razani et~al.(2021{\natexlab{b}})Razani, Cheng, Taghavi, and Bingbing]{razani2021litehdseg}
Ryan Razani, Ran Cheng, Ehsan Taghavi, and Liu Bingbing.
\newblock {Lite-HDSeg: LiDAR Semantic Segmentation Using Lite Harmonic Dense Convolutions}.
\newblock \emph{arXiv preprint arXiv:2103.08852}, 2021{\natexlab{b}}.

\bibitem[Ren et~al.(2024)Ren, Liu, Zeng, Lin, Li, Cao, Chen, Huang, Chen, Yan, Zeng, Zhang, Li, Yang, Li, Jiang, and Zhang]{GroundedSAM}
Tianhe Ren, Shilong Liu, Ailing Zeng, Jing Lin, Kunchang Li, He Cao, Jiayu Chen, Xinyu Huang, Yukang Chen, Feng Yan, Zhaoyang Zeng, Hao Zhang, Feng Li, Jie Yang, Hongyang Li, Qing Jiang, and Lei Zhang.
\newblock Grounded sam: Assembling open-world models for diverse visual tasks, 2024.

\bibitem[Ronneberger et~al.(2015)Ronneberger, Fischer, and Brox]{UNet}
Olaf Ronneberger, Philipp Fischer, and Thomas Brox.
\newblock U-net: Convolutional networks for biomedical image segmentation.
\newblock \emph{CoRR}, abs/1505.04597, 2015.

\bibitem[Ryali et~al.(2023)Ryali, Hu, Bolya, Wei, Fan, Huang, Aggarwal, Chowdhury, Poursaeed, Hoffman, Malik, Li, and Feichtenhofer]{Hiera}
Chaitanya Ryali, Yuan-Ting Hu, Daniel Bolya, Chen Wei, Haoqi Fan, Po-Yao Huang, Vaibhav Aggarwal, Arkabandhu Chowdhury, Omid Poursaeed, Judy Hoffman, Jitendra Malik, Yanghao Li, and Christoph Feichtenhofer.
\newblock Hiera: a hierarchical vision transformer without the bells-and-whistles.
\newblock In \emph{Proceedings of the 40th International Conference on Machine Learning}. JMLR.org, 2023.

\bibitem[Shahraki et~al.(2025)Shahraki, Elamin, and El-Rabbany]{samNetPlusPlus}
Mohsen Shahraki, Ahmed Elamin, and Ahmed El-Rabbany.
\newblock Samnet++: A segment anything model for supervised 3d point cloud semantic segmentation.
\newblock \emph{Remote Sensing}, 17\penalty0 (7), 2025.

\bibitem[Siméoni et~al.(2025)Siméoni, Vo, Seitzer, Baldassarre, Oquab, Jose, Khalidov, Szafraniec, Yi, Ramamonjisoa, Massa, Haziza, Wehrstedt, Wang, Darcet, Moutakanni, Sentana, Roberts, Vedaldi, Tolan, Brandt, Couprie, Mairal, Jégou, Labatut, and Bojanowski]{DINOv3}
Oriane Siméoni, Huy~V. Vo, Maximilian Seitzer, Federico Baldassarre, Maxime Oquab, Cijo Jose, Vasil Khalidov, Marc Szafraniec, Seungeun Yi, Michaël Ramamonjisoa, Francisco Massa, Daniel Haziza, Luca Wehrstedt, Jianyuan Wang, Timothée Darcet, Théo Moutakanni, Leonel Sentana, Claire Roberts, Andrea Vedaldi, Jamie Tolan, John Brandt, Camille Couprie, Julien Mairal, Hervé Jégou, Patrick Labatut, and Piotr Bojanowski.
\newblock Dinov3, 2025.

\bibitem[Sudre et~al.(2017)Sudre, Li, Vercauteren, Ourselin, and Cardoso]{diceLoss}
Carole~H. Sudre, Wenqi Li, Tom Vercauteren, S{\'{e}}bastien Ourselin, and M.~Jorge Cardoso.
\newblock Generalised dice overlap as a deep learning loss function for highly unbalanced segmentations.
\newblock \emph{CoRR}, abs/1707.03237, 2017.

\bibitem[Sun et~al.(2020)Sun, Kretzschmar, Dotiwalla, Chouard, Patnaik, Tsui, Guo, Zhou, Chai, Caine, Vasudevan, Han, Ngiam, Zhao, Timofeev, Ettinger, Krivokon, Gao, Joshi, Zhang, Shlens, Chen, and Anguelov]{waymo}
Pei Sun, Henrik Kretzschmar, Xerxes Dotiwalla, Aurelien Chouard, Vijaysai Patnaik, Paul Tsui, James Guo, Yin Zhou, Yuning Chai, Benjamin Caine, Vijay Vasudevan, Wei Han, Jiquan Ngiam, Hang Zhao, Aleksei Timofeev, Scott Ettinger, Maxim Krivokon, Amy Gao, Aditya Joshi, Yu Zhang, Jonathon Shlens, Zhifeng Chen, and Dragomir Anguelov.
\newblock Scalability in perception for autonomous driving: Waymo open dataset.
\newblock In \emph{Proceedings of the IEEE/CVF Conference on Computer Vision and Pattern Recognition (CVPR)}, 2020.

\bibitem[Sun et~al.(2023{\natexlab{a}})Sun, Fang, Wu, Wang, and Cao]{EVA-CLIP}
Quan Sun, Yuxin Fang, Ledell Wu, Xinlong Wang, and Yue Cao.
\newblock Eva-clip: Improved training techniques for clip at scale.
\newblock \emph{arXiv preprint arXiv:2303.15389}, 2023{\natexlab{a}}.

\bibitem[Sun et~al.(2023{\natexlab{b}})Sun, Wang, Yu, Cui, Zhang, Zhang, and Wang]{EVA-CLIP-18B}
Quan Sun, Jinsheng Wang, Qiying Yu, Yufeng Cui, Fan Zhang, Xiaosong Zhang, and Xinlong Wang.
\newblock Eva-clip-18b: Scaling clip to 18 billion parameters.
\newblock \emph{arXiv preprint arXiv:2402.04252}, 2023{\natexlab{b}}.

\bibitem[Tan et~al.(2023)Tan, Dong, Zhang, Zhang, Ji, and Li]{ovoopenvocabularyoccupancy}
Zhiyu Tan, Zichao Dong, Cheng Zhang, Weikun Zhang, Hang Ji, and Hao Li.
\newblock Ovo: Open-vocabulary occupancy, 2023.

\bibitem[Tang et~al.(2020)Tang, Liu, Zhao, Lin, Lin, Wang, and Han]{SPVNAS}
Haotian Tang, Zhijian Liu, Shengyu Zhao, Yujun Lin, Ji Lin, Hanrui Wang, and Song Han.
\newblock Searching efficient 3d architectures with sparse point-voxel convolution.
\newblock \emph{CoRR}, abs/2007.16100, 2020.

\bibitem[Thomas et~al.(2019)Thomas, Qi, Deschaud, Marcotegui, Goulette, and Guibas]{KPConvFA}
Hugues Thomas, C. Qi, Jean-Emmanuel Deschaud, Beatriz Marcotegui, François Goulette, and Leonidas~J. Guibas.
\newblock Kpconv: Flexible and deformable convolution for point clouds.
\newblock \emph{2019 IEEE/CVF International Conference on Computer Vision (ICCV)}, pages 6410--6419, 2019.

\bibitem[Triess et~al.(2020)Triess, Peter, Rist, and Zöllner]{ScanBasedSeg}
Larissa~T. Triess, David Peter, Christoph~B. Rist, and J.~Marius Zöllner.
\newblock Scan-based semantic segmentation of lidar point clouds: An experimental study.
\newblock In \emph{2020 IEEE Intelligent Vehicles Symposium (IV)}, pages 1116--1121, 2020.

\bibitem[Vaswani et~al.(2017)Vaswani, Shazeer, Parmar, Uszkoreit, Jones, Gomez, Kaiser, and Polosukhin]{attention}
Ashish Vaswani, Noam Shazeer, Niki Parmar, Jakob Uszkoreit, Llion Jones, Aidan~N. Gomez, Lukasz Kaiser, and Illia Polosukhin.
\newblock Attention is all you need.
\newblock \emph{CoRR}, abs/1706.03762, 2017.

\bibitem[Veličković et~al.(2018)Veličković, Cucurull, Casanova, Romero, Liò, and Bengio]{graphattentionnetworks}
Petar Veličković, Guillem Cucurull, Arantxa Casanova, Adriana Romero, Pietro Liò, and Yoshua Bengio.
\newblock Graph attention networks, 2018.

\bibitem[Wang et~al.(2019)Wang, Wu, Wu, and Keutzer]{wang2019latte}
Bernie Wang, Virginia Wu, Bichen Wu, and Kurt Keutzer.
\newblock Latte: accelerating lidar point cloud annotation via sensor fusion, one-click annotation, and tracking.
\newblock In \emph{2019 IEEE Intelligent Transportation Systems Conference (ITSC)}, pages 265--272. IEEE, 2019.

\bibitem[Wang et~al.(2022)Wang, Zhu, and Zhang]{MetaRangeSeg}
Song Wang, Jianke Zhu, and Ruixiang Zhang.
\newblock Meta-rangeseg: Lidar sequence semantic segmentation using multiple feature aggregation.
\newblock \emph{IEEE Robotics and Automation Letters}, 7\penalty0 (4):\penalty0 9739–9746, 2022.

\bibitem[Wang et~al.(2018)Wang, Sun, Liu, Sarma, Bronstein, and Solomon]{DynGraphCNN}
Yue Wang, Yongbin Sun, Ziwei Liu, Sanjay~E. Sarma, Michael~M. Bronstein, and Justin~M. Solomon.
\newblock Dynamic graph {CNN} for learning on point clouds.
\newblock \emph{CoRR}, abs/1801.07829, 2018.

\bibitem[Wang et~al.(2024)Wang, Ning, and Blaschko]{wang2024jaccardmetriclossesoptimizing}
Zifu Wang, Xuefei Ning, and Matthew~B. Blaschko.
\newblock Jaccard metric losses: Optimizing the jaccard index with soft labels, 2024.

\bibitem[Wu et~al.(2018)Wu, Wan, Yue, and Keutzer]{wu2017squeezeseg}
Bichen Wu, Alvin Wan, Xiangyu Yue, and Kurt Keutzer.
\newblock Squeezeseg: Convolutional neural nets with recurrent crf for real-time road-object segmentation from 3d lidar point cloud.
\newblock In \emph{ICRA}, 2018.

\bibitem[Wu et~al.(2019{\natexlab{a}})Wu, Zhou, Zhao, Yue, and Keutzer]{wu2018squeezesegv2}
Bichen Wu, Xuanyu Zhou, Sicheng Zhao, Xiangyu Yue, and Kurt Keutzer.
\newblock Squeezesegv2: Improved model structure and unsupervised domain adaptation for road-object segmentation from a lidar point cloud.
\newblock In \emph{ICRA}, 2019{\natexlab{a}}.

\bibitem[Wu et~al.(2023)Wu, Guo, Li, Yu, Gao, and Sang]{boundaryLoss}
Dongyue Wu, Zilin Guo, Aoyan Li, Changqian Yu, Changxin Gao, and Nong Sang.
\newblock Conditional boundary loss for semantic segmentation.
\newblock \emph{IEEE Transactions on Image Processing}, 32:\penalty0 3717--3731, 2023.

\bibitem[Wu et~al.(2019{\natexlab{b}})Wu, Qi, and Fuxin]{PointConv}
Wenxuan Wu, Zhongang Qi, and Li Fuxin.
\newblock Pointconv: Deep convolutional networks on 3d point clouds.
\newblock In \emph{2019 IEEE/CVF Conference on Computer Vision and Pattern Recognition (CVPR)}, pages 9613--9622, 2019{\natexlab{b}}.

\bibitem[Wu et~al.(2022)Wu, Lao, Jiang, Liu, and Zhao]{ptv2}
Xiaoyang Wu, Yixing Lao, Li Jiang, Xihui Liu, and Hengshuang Zhao.
\newblock Point transformer v2: Grouped vector attention and partition-based pooling.
\newblock In \emph{NeurIPS}, 2022.

\bibitem[Wu et~al.(2024{\natexlab{a}})Wu, Jiang, Wang, Liu, Liu, Qiao, Ouyang, He, and Zhao]{ptv3}
Xiaoyang Wu, Li Jiang, Peng-Shuai Wang, Zhijian Liu, Xihui Liu, Yu Qiao, Wanli Ouyang, Tong He, and Hengshuang Zhao.
\newblock Point transformer v3: Simpler, faster, stronger.
\newblock In \emph{CVPR}, 2024{\natexlab{a}}.

\bibitem[Wu et~al.(2024{\natexlab{b}})Wu, Tian, Wen, Peng, Liu, Yu, and Zhao]{ppt}
Xiaoyang Wu, Zhuotao Tian, Xin Wen, Bohao Peng, Xihui Liu, Kaicheng Yu, and Hengshuang Zhao.
\newblock Towards large-scale 3d representation learning with multi-dataset point prompt training.
\newblock In \emph{CVPR}, 2024{\natexlab{b}}.

\bibitem[Xiao et~al.(2023)Xiao, Wu, Xu, Dai, Hu, Lu, Zeng, Liu, and Yuan]{florence2advancingunifiedrepresentation}
Bin Xiao, Haiping Wu, Weijian Xu, Xiyang Dai, Houdong Hu, Yumao Lu, Michael Zeng, Ce Liu, and Lu Yuan.
\newblock Florence-2: Advancing a unified representation for a variety of vision tasks, 2023.

\bibitem[Xiong et~al.(2024)Xiong, Wu, Tan, Li, Tang, Chen, Li, Ma, and Li]{sam2unet}
Xinyu Xiong, Zihuang Wu, Shuangyi Tan, Wenxue Li, Feilong Tang, Ying Chen, Siying Li, Jie Ma, and Guanbin Li.
\newblock Sam2-unet: Segment anything 2 makes strong encoder for natural and medical image segmentation.
\newblock \emph{arXiv preprint arXiv:2408.08870}, 2024.

\bibitem[Xu et~al.(2020)Xu, Wu, Wang, Zhan, Vajda, Keutzer, and Tomizuka]{xu2020squeezesegv3}
Chenfeng Xu, Bichen Wu, Zining Wang, Wei Zhan, Peter Vajda, Kurt Keutzer, and Masayoshi Tomizuka.
\newblock Squeezesegv3: Spatially-adaptive convolution for efficient point-cloud segmentation.
\newblock In \emph{European Conference on Computer Vision}, pages 1--19. Springer, 2020.

\bibitem[Xu et~al.(2021)Xu, Wu, Zhan, Vajda, Keutzer, and Tomizuka]{xu2021rangevit}
Chenfeng Xu, Bichen Wu, Wei Zhan, Peter Vajda, Kurt Keutzer, and Masayoshi Tomizuka.
\newblock Rangevit: Vision transformer for lidar range image segmentation.
\newblock In \emph{Proceedings of the IEEE/CVF International Conference on Computer Vision (ICCV)}, pages 27069--27078, 2021.

\bibitem[Xu et~al.(2025{\natexlab{a}})Xu, Wang, Ni, Hu, Yang, Zhu, and Li]{sam4d}
Jianyun Xu, Song Wang, Ziqian Ni, Chunyong Hu, Sheng Yang, Jianke Zhu, and Qiang Li.
\newblock Sam4d: Segment anything in camera and lidar streams, 2025{\natexlab{a}}.

\bibitem[Xu et~al.(2025{\natexlab{b}})Xu, Kong, Shuai, and Liu]{xu2025frnet}
Xiang Xu, Lingdong Kong, Hui Shuai, and Qingshan Liu.
\newblock Frnet: Frustum-range networks for scalable lidar segmentation.
\newblock \emph{IEEE Transactions on Image Processing}, 34:\penalty0 2173--2186, 2025{\natexlab{b}}.

\bibitem[Yan et~al.(2020)Yan, Zheng, Li, Wang, and Cui]{PointASNL}
Xu Yan, Chaoda Zheng, Zhen Li, Sheng Wang, and Shuguang Cui.
\newblock Pointasnl: Robust point clouds processing using nonlocal neural networks with adaptive sampling.
\newblock In \emph{2020 IEEE/CVF Conference on Computer Vision and Pattern Recognition (CVPR)}, pages 5588--5597, 2020.

\bibitem[Yang et~al.(2023)Yang, Wu, He, Zhao, and Liu]{sam3dsegment3dscenes}
Yunhan Yang, Xiaoyang Wu, Tong He, Hengshuang Zhao, and Xihui Liu.
\newblock Sam3d: Segment anything in 3d scenes, 2023.

\bibitem[Yuan et~al.(2025)Yuan, Li, Zhang, Huang, Xu, Ji, Tong, Qi, Feng, and Yang]{yuan2025sa2vamarryingsam2llava}
Haobo Yuan, Xiangtai Li, Tao Zhang, Zilong Huang, Shilin Xu, Shunping Ji, Yunhai Tong, Lu Qi, Jiashi Feng, and Ming-Hsuan Yang.
\newblock Sa2va: Marrying sam2 with llava for dense grounded understanding of images and videos, 2025.

\bibitem[Yuan et~al.(2021)Yuan, Chen, Chen, Codella, Dai, Gao, Hu, Huang, Li, Li, Liu, Liu, Liu, Lu, Shi, Wang, Wang, Xiao, Xiao, Yang, Zeng, Zhou, and Zhang]{florence1}
Lu Yuan, Dongdong Chen, Yi{-}Ling Chen, Noel Codella, Xiyang Dai, Jianfeng Gao, Houdong Hu, Xuedong Huang, Boxin Li, Chunyuan Li, Ce Liu, Mengchen Liu, Zicheng Liu, Yumao Lu, Yu Shi, Lijuan Wang, Jianfeng Wang, Bin Xiao, Zhen Xiao, Jianwei Yang, Michael Zeng, Luowei Zhou, and Pengchuan Zhang.
\newblock Florence: {A} new foundation model for computer vision.
\newblock \emph{CoRR}, abs/2111.11432, 2021.

\bibitem[Zeid et~al.(2025)Zeid, Yilmaz, de~Geus, Hermans, Adrian, Linder, and Leibe]{dinoroomleveraging2d}
Karim~Abou Zeid, Kadir Yilmaz, Daan de Geus, Alexander Hermans, David Adrian, Timm Linder, and Bastian Leibe.
\newblock Dino in the room: Leveraging 2d foundation models for 3d segmentation, 2025.

\bibitem[Zhang et~al.(2023)Zhang, Dong, and Ma]{CLIPFO3D}
Junbo Zhang, Runpei Dong, and Kaisheng Ma.
\newblock Clip-fo3d: Learning free open-world 3d scene representations from 2d dense clip.
\newblock In \emph{2023 IEEE/CVF International Conference on Computer Vision Workshops (ICCVW)}, pages 2040--2051, 2023.

\bibitem[Zhang et~al.(2020)Zhang, Zhou, David, Yue, Xi, Gong, and Foroosh]{PolarNet}
Yang Zhang, Zixiang Zhou, Philip David, Xiangyu Yue, Zerong Xi, Boqing Gong, and Hassan Foroosh.
\newblock Polarnet: An improved grid representation for online lidar point clouds semantic segmentation.
\newblock In \emph{2020 IEEE/CVF Conference on Computer Vision and Pattern Recognition (CVPR)}, pages 9598--9607, 2020.

\bibitem[Zhang et~al.(2022)Zhang, Zhou, Foroosh, Lauri, Li, Xu, and Lepetit]{zhang2022rangeformer}
Zhen Zhang, Yin Zhou, Hassan Foroosh, Mikko Lauri, Shiguang Li, Hongkai Xu, and Vincent Lepetit.
\newblock Rangeformer: Point cloud semantic segmentation using range image based transformers.
\newblock In \emph{Advances in Neural Information Processing Systems (NeurIPS)}, pages 22397--22410, 2022.

\bibitem[Zhao et~al.(2025)Zhao, Wang, Kannala, and Pajarinen]{zhao2025multiscalefusionobjectrepresentation}
Rongzhen Zhao, Vivienne Wang, Juho Kannala, and Joni Pajarinen.
\newblock Multi-scale fusion for object representation, 2025.

\bibitem[Zhao et~al.(2021)Zhao, Bai, and Huang]{zhao2021fidnet}
Yiming Zhao, Lin Bai, and Xinming Huang.
\newblock {FIDNet: LiDAR Point Cloud Semantic Segmentation with Fully Interpolation Decoding}.
\newblock In \emph{IEEE/RSJ International Conference on Intelligent Robots and Systems (IROS)}, 2021.

\bibitem[Zhou et~al.(2020)Zhou, Zhu, Song, Ma, Wang, Li, and Lin]{Cylinder3D}
Hui Zhou, Xinge Zhu, Xiao Song, Yuexin Ma, Zhe Wang, Hongsheng Li, and Dahua Lin.
\newblock Cylinder3d: An effective 3d framework for driving-scene lidar semantic segmentation.
\newblock \emph{CoRR}, abs/2008.01550, 2020.

\bibitem[Zhuang et~al.(2021)Zhuang, Li, Jia, Wang, Li, and Tan]{PerceptionAware}
Zhuangwei Zhuang, Rong Li, Kui Jia, Qicheng Wang, Yuanqing Li, and Mingkui Tan.
\newblock Perception-aware multi-sensor fusion for 3d lidar semantic segmentation.
\newblock In \emph{2021 IEEE/CVF International Conference on Computer Vision (ICCV)}, pages 16260--16270, 2021.

\bibitem[Zou et~al.(2023)Zou, Yang, Zhang, Li, Li, Wang, Wang, Gao, and Lee]{seem}
Xueyan Zou, Jianwei Yang, Hao Zhang, Feng Li, Linjie Li, Jianfeng Wang, Lijuan Wang, Jianfeng Gao, and Yong~Jae Lee.
\newblock Segment everything everywhere all at once, 2023.

\bibitem[Özdemir and Sönmez(2020)]{WCELoss}
Özgür Özdemir and Elena~Battini Sönmez.
\newblock Weighted cross-entropy for unbalanced data with application on covid x-ray images.
\newblock In \emph{2020 Innovations in Intelligent Systems and Applications Conference (ASYU)}, pages 1--6, 2020.

\end{thebibliography}
}

\end{document}